\documentclass[sigconf]{acmart}

\AtBeginDocument{%
  }

\setcopyright{acmlicensed}
\copyrightyear{2026}
\acmYear{2026}
\setcopyright{cc}
\setcctype{by}
\acmConference[CIKM '26]{Proceedings of the 35th ACM International Conference on Information and Knowledge Management}{November 07--11, 2026}{Rome, Italy}
\acmBooktitle{Proceedings of the 35th ACM International Conference on Information and Knowledge Management (CIKM '26), November 07--11, 2026, Rome, Italy}
\acmDOI{10.1145/3799682.3841155}
\acmISBN{979-8-4007-2539-5/2026/11}

\usepackage{amsmath}
\usepackage{amsthm}
\usepackage{array}
\usepackage{adjustbox}
\usepackage{algorithm}
\usepackage{algorithmic}
\usepackage{booktabs}
\usepackage{caption}
\usepackage{colortbl}
\usepackage{enumitem}
\usepackage{graphicx}
\usepackage{makecell}
\usepackage{multirow}
\usepackage{stfloats}
\usepackage{xcolor}
\usepackage{hyperref}

\definecolor{darkred}{RGB}{178,34,34}

\graphicspath{{Figures/}}

\ccsdesc[500]{Information systems~Multimedia information systems}
\keywords{Remote Sensing, Contrastive Learning, Urban Indicator Prediction}

\title{CoST: Semantic-Aware Urban Understanding via Spatial-Temporal Alignment}

\author{Yutian Jiang}
\authornote{Yutian Jiang and Jiabo Liu contributed equally to this research.}
\email{yjiang194@connect.hkust-gz.edu.cn}
\affiliation{%
  \institution{The Hong Kong University of Science and Technology (Guangzhou)}
  \city{Guangzhou}
  \country{China}}

\author{Jiabo Liu}
\authornotemark[1]
\email{jliu933@connect.hkust-gz.edu.cn}
\affiliation{%
  \institution{The Hong Kong University of Science and Technology (Guangzhou)}
  \city{Guangzhou}
  \country{China}}

\author{Xixuan Hao}
\email{xhao390@connect.hkust-gz.edu.cn}
\affiliation{%
  \institution{The Hong Kong University of Science and Technology (Guangzhou)}
  \city{Guangzhou}
  \country{China}}

\author{Yuxuan Liang}
\correspondingauthor
\email{yuxliang@outlook.com}
\affiliation{%
  \institution{The Hong Kong University of Science and Technology (Guangzhou)}
  \city{Guangzhou}
  \country{China}}

\renewcommand{\shortauthors}{Jiang, Liu, Hao, and Liang}

\begin{document}

\begin{abstract}

Geospatial representation learning from satellite imagery is a fundamental problem for large-scale urban analysis and real-world applications. Despite recent advances, current methods struggle with cross-region generalization and semantic interpretability due to their reliance on region-specific auxiliary data and the neglect of semantic alignment within multi-temporal urban imagery. Therefore, we present CoST, a novel \underline{Co}ntrastive-based \underline{S}patial-\underline{T}emporal framework that aligns spatial context with multi-temporal semantics to extract universal geographic regularities shared across regions. Specifically, CoST explicitly models spatial correlations to capture transferable geographic structures and exploits multi-year urban change semantics to align learned representations with high-level geo-semantics. Extensive experiments demonstrate that CoST consistently achieves superior performance across various downstream tasks and in unseen scenario, yielding an average relative gain of 8.7\% over the strongest competing methods across eight city-indicator settings. The code is available in \href{https://github.com/Arandinglv/CoST}{this repo}.
\end{abstract}

\maketitle

\section{Introduction}
\begin{figure}
    \centering
    \includegraphics[width=\linewidth]{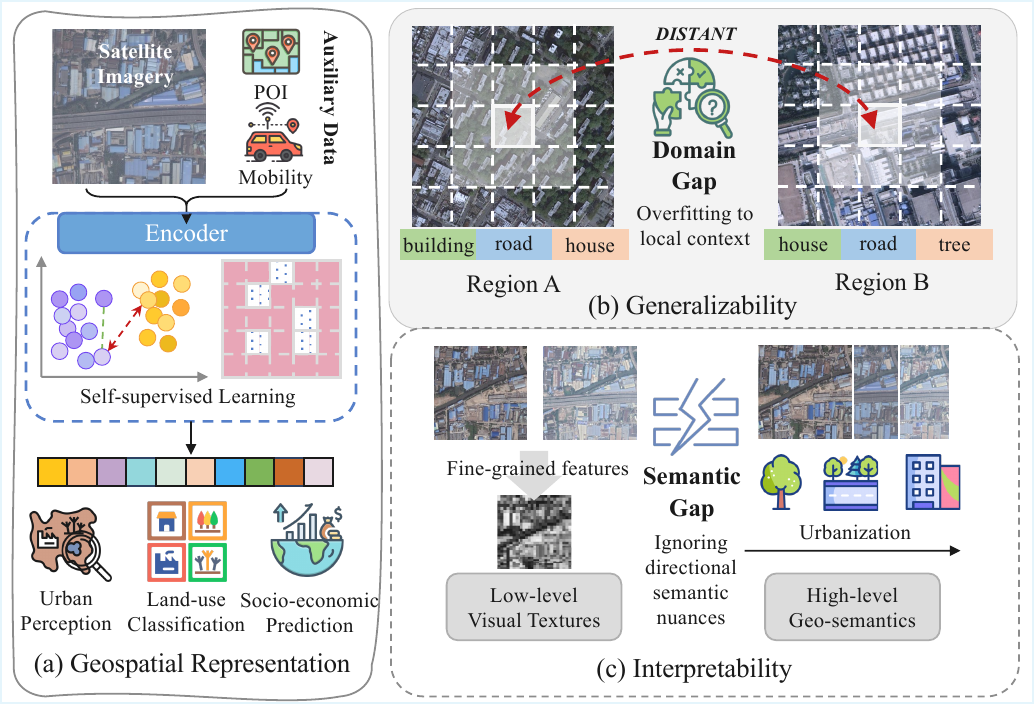}
    \caption{Geospatial representation and its challenges.}
    \label{fig:definition}
\end{figure}

Geospatial representation learning aims to project multi-source data into a unified latent space, effectively capturing intrinsic spatial correlations and temporal dynamics to facilitate robust urban understanding~\cite{wu2024torchspatial}. As illustrated in Figure~\ref{fig:definition}~(a), a meaningful geospatial representation should capture not only local visual patterns but also the higher-level geographic regularities that shape urban structure and its evolution over time~\cite{hao2025geospatial}. 
The learned geospatial representations can support a wide range of downstream applications, such as urban perception and socio-economic assessment \cite{urbanr1}. Moreover, the quality of these representations fundamentally determines their ability to generalize and transfer across diverse geographic regions and tasks. Therefore, learning transferable and semantically meaningful geospatial representations has become a core problem in data-driven urban system understanding \cite{urbancomputing}.

Recently, satellite imagery has emerged as a primary modality for geospatial representation learning that provides large-scale, geographically consistent observations of the Earth's surface~\cite{hao2024urbanvlp}. Existing research on satellite-based representation learning generally follows two paradigms: (1) \textit{Multimodal Semantic Alignment}, which incorporates socio-economic semantics by aligning satellite imagery with auxiliary data, such as POIs \cite{sun2025flexireg,regionencoder} or mobility data \cite{zhao2025graphjcl,yong2024musecl}. While effective for region-specific profiling and real-world socio-economic prediction, these methods depend heavily on external data sources whose availability, quality, and distribution vary substantially across cities and countries, which limits scalability and weakens transferability. (2) \textit{Self-supervised Visual Pre-training}, which focuses on extracting visual features by mining invariant textural patterns directly from raw imagery. Although such methods produce strong visual encoders for low-level tasks, they primarily emphasize appearance consistency and texture discrimination, with limited explicit grounding in high-level geo-semantics. As a result, the learned features are often effective as visual descriptors but less informative for semantically demanding urban analysis. 

These limitations are not independent, pointing to a more fundamental gap in current geospatial representation learning: a representation should generalize across regions with different visual styles, development patterns, and functional compositions, while also remaining semantically interpretable enough to reflect meaningful urban transitions. This leads to two central challenges in learning universal geospatial representations: (i) generalization under cross-region distribution shifts, and (ii) semantic interpretability under weak supervision.

\begin{itemize}[leftmargin=*]
    \item \textbf{Cross-region Generalization}. Cross-region transfer remains a key challenge in urban representation learning~\cite{urbanagent,hao2024urbanvlp,zhong2024urbancross}. Methods relying on auxiliary signals, such as points of interest (POIs) or mobility data, suffer from limited generalization due to substantial distributional discrepancies across cities, causing learned representations to become entangled with region-specific characteristics. A model trained in one source city may therefore suffer severe performance degradation when transferred to visually and functionally dissimilar cities, or when deployed on downstream tasks that differ from those seen during training. As shown in Figure~\ref{fig:definition}~(b), models trained on dense metropolitan environments such as New York City may struggle to generalize to suburban regions because their representations become biased toward city-specific distributions. Moreover, current multimodal approaches require data recollection and model retraining for each new region~\cite{fang2022transfer}, highlighting the need for representations that capture transferable geographic structures rather than region-specific correlations.

    \item \textbf{Semantic Interpretability}. Multi-temporal satellite imagery naturally contains rich semantic transition cues across time and space. Such transitions vary across locations (e.g., \textit{land $\rightarrow$ road} versus \textit{land $\rightarrow$ building}) and exhibit temporal continuity at the same location, such as \textit{land $\rightarrow$ road $\rightarrow$ building}. Modeling these dynamic visual patterns together with explicit geo-semantics is therefore critical for learning interpretable geospatial representations. Yet this remains difficult because dense semantic annotations over long temporal horizons are scarce. Although prior work has incorporated temporal information, most existing methods reduce temporal variation to a binary change/no-change signal, which provides only weak supervision and fails to characterize the underlying semantic transitions. As a result, the learned representations are ambiguous and lack semantic interpretability, as illustrated in Figure~\ref{fig:definition}(c).
    
\end{itemize}

To bridge the gap, we propose \textbf{CoST}, a novel \underline{Co}ntrastive-based \underline{S}patial-\underline{T}emporal framework for geospatial representation, which couples space and time for cross-region generalization and semantic interpretability. The key idea is that robust urban representations should be shaped by local spatial proximity and temporal consistency of semantic transitions in neighboring regions. To improve cross-region generalization, CoST introduces a spatial neighborhood modeling strategy inspired by Tobler's First Law of Geography~\cite{tobler1970computer}, which posits that geographic proximity implies semantic similarity. This strategy preserves local continuity while remaining sensitive to instance discrimination, enabling the model to capture invariant spatial structures shared across diverse urban environments. To enhance semantic interpretability, CoST treats multi-temporal imagery as a semantic pretext through encoding temporal dynamics with embeddings that capture both the \textit{semantic type} and \textit{change extent} of transitions, providing richer semantic supervision than binary change detection and enabling more interpretable representations. Finally, CoST introduces a spatial-temporal alignment mechanism that enforces consistency among neighboring regions exhibiting similar change trajectories, thereby linking spatial similarity with temporal semantics and improving the coherence of the learned representation space.

Overall, the contributions can be summarized as follows: 
\begin{itemize}[leftmargin=*, itemsep=0pt]
    \item  
    We propose {CoST}, a unified framework for learning geospatial representations from satellite imagery by jointly modeling spatial neighborhood structure and temporal semantic transitions, with the goal of improving both cross-region generalization and semantic interpretability.
    
    \item We introduce three complementary components: \textit{spatial neighborhood modeling} to capture transferable local geographic structure, \textit{temporal semantic guidance} to encode multi-year semantic transitions with both change type and change extent, and \textit{spatial-temporal alignment} to couple spatial proximity with temporal consistency in a shared latent space.
    
    \item We curate a spatial-temporal aligned satellite imagery dataset spanning 11 years and multiple major cities, and conduct extensive experiments across cross-city and cross-task settings. The results show that CoST consistently outperforms strong baselines and yields more semantically structured embeddings for urban analysis.

\end{itemize}

\section{Related Works}
\noindent \textbf{Multimodal Alignment.}
This line of research learns urban or regional representations by injecting external semantic knowledge into satellite imagery through feature fusion or cross-modal alignment~\cite{zou2025deep}. The core idea is that auxiliary modalities such as POIs, human mobility, and textual descriptions provide complementary semantics about land use, human activity, and urban dynamics~\cite{mvgrnet}. Representative studies align satellite imagery with POI semantics~\cite{liu2023knowcl,xiao2024refound}, mobility patterns~\cite{urbanincontext,liu2024online}, or graph-structured urban context~\cite{wen2024demo2vec,cai2024explainable,geohg} to learn semantically enriched region embeddings. More recent methods further adopt contrastive alignment objectives to associate satellite imagery with textual or functional descriptions, improving region profiling and urban indicator prediction~\cite{yan2024urbanclip,USPM, muhlematter2025urbanfusion}. Despite their strong performance, these methods fundamentally depend on auxiliary data sources whose coverage and quality vary substantially across cities. As a result, their learned representations are entangled with region-specific external knowledge, which limits scalability and weakens transferability to new regions where such signals are sparse, noisy, or unavailable.

\noindent \textbf{Visual Pre-training.}
Visual pre-training aims to learn generic representations directly from images. Recent geospatial foundation models exploit self-supervised objectives such as contrastive learning and masked image modeling to capture invariant visual patterns from large-scale image data. Contrastive approaches learn instance discrimination and augmentation invariance from raw imagery, while Masked Autoencoder (MAE) based methods reconstruct missing image patches to model local texture and spatial structure. Representative examples include SatMAE~\cite{cong2022satmae}, SatMAE++~\cite{satmae++}, and ScaleMAE~\cite{reed2023scalemae}, which have shown strong transferability across land-usage classification or segmentation tasks. However, their training signals are usually defined by visual consistency within imagery itself, without explicitly constraining the representation space to reflect urban semantics. Despite learning powerful generic visual features, these methods typically produce embeddings that are less structured for high-level urban understanding and socio-economic inference.

\noindent \textbf{Spatial-Temporal Modeling.}
Temporal imagery provides a natural signal for modeling the inherent dynamics of urban regions. Contrastive learning approaches \cite{manas2021seco,mall2023caco} learn temporal invariance by treating observations from different dates as related views, improving robustness for downstream visual tasks, and MAE-based methods \cite{guo2024skysense} incorporate multi-temporal observations to enhance large-scale visual understanding. Recent studies in geospatial foundation modeling, such as Prithvi \cite{prithvi_eo2}, also show that temporal stacks can substantially improve adaptation across different tasks by pre-training over multi-temporal satellite sequences. Nevertheless, most existing temporal methods still treat geospatial tiles as isolated instances, overlooking the coupling of spatial-temporal dimensions. This oversight causes a disconnect between low-level visual textures and high-level geospatial semantics, leading to a performance trade-off: models may excel at low-level vision tasks but struggle with socio-economic inference. Consequently, they fail to achieve a unified representation space that is robust for various downstream applications.

\section{Methodology}

\begin{figure*}[htbp]
    \centering
    \includegraphics[width=\textwidth]{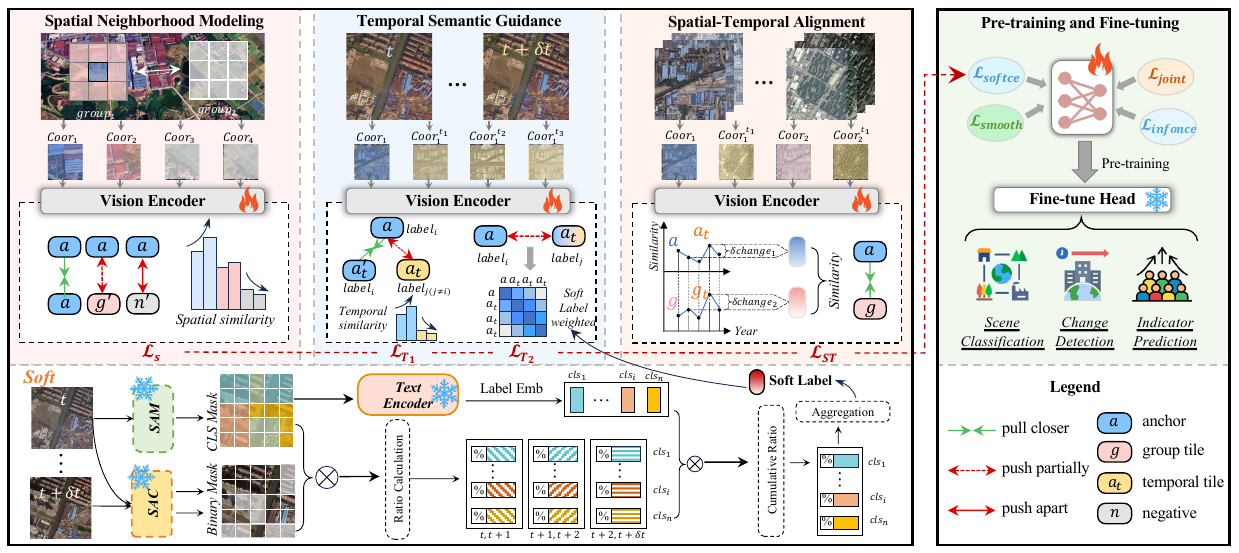} 
    \caption{Overall framework of the proposed CoST, aligning geographical neighborhood structures with multi-year urban change semantics to learn generalizable and interpretable geospatial representations for various urban analysis tasks.}
    \label{fig:framework}
\end{figure*}

\subsection{Preliminaries}

\noindent \textbf{Satellite Image.}
We formally define geospatial representation learning from satellite imagery. Let $\mathcal{D} = \{I_{i,t}\}$ denote a collection of satellite images, where $i \in \{1,\dots,N\}$ indexes discrete geospatial regions and $t \in \mathcal{T}=\{1,\dots,T\}$ indexes time. For a given region $i$, a satellite image observed at time $t$ is denoted as $I_{i,t} \in \mathbb{R}^{H \times W \times C}$, where $H$ and $W$ are the image height and width, and $C$ is the number of channels. Accordingly, the multi-temporal satellite observations of region $i$ are represented as a sequence $\mathcal{X}_i=\{I_{i,1},I_{i,2},\dots,I_{i,T}\}$ over the temporal period $\mathcal{T}$.

\noindent \textbf{Geospatial Regions.}
Each satellite tile is associated with a geographic coordinate $\mathbf{l}_i \in \mathbb{R}^{2}$. Based on geographic distance, we define the spatial neighborhood of region $i$ as $\mathcal{N}_i=\{j_1,j_2,\dots,j_K\}$, where each $j_k$ denotes one of its $K$ nearest neighboring regions. Therefore, each sample is characterized by both its temporal observations $\mathcal{X}_i$ and its spatial context $\mathcal{N}_i$. This formulation provides the basic spatial-temporal structure used in our framework.

\noindent \textbf{Problem Statement.}
Given a satellite tile $I_{i,t}$, the goal is to learn a generalizable visual encoder $\mathcal{E}_{\theta}$ that maps it to a representation vector $\mathbf{z}_{i,t}=\mathcal{E}_{\theta}(I_{i,t}) \in \mathbb{R}^{d}$. During pre-training, the encoder is optimized without downstream labels. After pre-training, the encoder is transferred to downstream tasks with frozen representations, and a lightweight task-specific head $\mathcal{H}_{\phi}$ is trained to produce predictions $\hat{y}_{i,t}=\mathcal{H}_{\phi}(\mathbf{z}_{i,t})$. For temporal downstream tasks, the head can further operate on the representation sequence $\mathbf{Z}_i=\{\mathbf{z}_{i,1},\dots,\mathbf{z}_{i,T}\}$. The objective is to learn transferable geospatial representations that support diverse downstream tasks across regions.

\subsection{Overview}

Figure~\ref{fig:framework} presents the overall paradigm of \textbf{CoST}, a unified spatial-temporal representation learning framework designed to learn geospatial embeddings that are both transferable across regions and semantically interpretable for urban analysis. Rather than modeling satellite tiles as isolated visual instances, CoST jointly exploits two inherent structures of urban observations: \textit{spatial dependency} across neighboring regions and \textit{temporal evolution} within the same region over time. To this end, the framework consists of three complementary components. First, \textit{spatial neighborhood modeling} captures local geographic dependency by encouraging semantically related neighboring regions to form consistent representations, thereby improving cross-region generalization. Second, \textit{temporal semantic guidance} introduces supervision from multi-temporal urban changes, enabling the encoder to distinguish not only whether a region changes, but also the semantic direction and magnitude of such change. Third, \textit{spatial-temporal alignment} couples the above two views by constraining neighboring regions with similar temporal transition patterns to remain aligned in the latent space, yielding a more coherent representation structure. After pre-training, the learned encoder is transferred to multiple downstream tasks with lightweight task-specific heads, including socio-economic indicator prediction, land-use classification, and change detection.
 
\subsection{Spatial Neighborhood Modeling}
\label{spatial_neighborhood}

Standard contrastive learning treats regions as independent instances, ignoring that geographically proximate areas share underlying environmental contexts~\cite{satcle}. To improve generalization across heterogeneous landscapes, we explicitly model spatial structure by aggregating nearby regions into coherent neighborhoods, following Tobler’s First Law of Geography \cite{tobler1970computer}. Formally, for each anchor $I_i$, we define a spatial neighborhood $\mathcal{N}_i = \{I_{{n}_{1}},I_{{n}_{2}} \dots, I_{{n}_{k}}\}$ consisting of its $k$ nearest geographic neighbors. Within a batch, we consider two types of positive signals: the instance-level positive $i^{+}$ and neighborhood-level positives drawn from $\mathcal{N}_i$. To incorporate this hierarchical structure into the representation space, we construct a target distribution $\mathcal{Q}_{i}$ to serve as a supervisory signal. For each sample $j$, we assign a target score $q_{i,j}$ that reflects its spatial proximity to the anchor $I_i$:

\begin{equation}
q_{i, j} =
\begin{cases}1, & \text{if } j = i^{+}  \\
\epsilon, & \text{if } j \in \mathcal{N}_i\\
0, & \text{otherwise}
\end{cases}, 
\label{eq:distribution}
\end{equation}
where $i^{+}$ denotes the augmented view of the anchor, and $\epsilon \in (0, 1)$ modulates the strength of spatial supervision. This distribution reflects a hierarchy in which the anchor itself provides the primary learning signal, while spatial neighbors offer auxiliary supervision to capture shared contexts. Instead of using a one-hot target, the model is trained to align its predicted similarity distribution with this target distribution $\mathcal{Q}_{i}$ by minimizing the following spatial loss: 
\begin{equation}
\mathcal{L}_s = - \sum_{j \in \mathcal{B}} q_{i, j} \cdot \log \frac{\exp(\mathbf{z}_i \cdot \mathbf{z}_j / \tau)}{\sum_{k \in \mathcal{B}} \exp(\mathbf{z}_i \cdot \mathbf{z}_k / \tau)}, 
\label{l_s}
\end{equation}
where $\mathcal{B}$ denotes the batch, $\mathbf{z}$ represents the normalized embedding, and $\tau$ is the temperature parameter. This objective encourages geographically proximate regions to be embedded nearby, preserving shared spatial semantics and improving transferability to unseen regions.

\subsection{Temporal Semantic Guidance}

Visual representations learned solely from pixel-level reconstruction or discrimination lack alignment with high-level geo-semantics because they prioritize low-level textures over functional meaning~\cite{interpretable}. This limitation is pronounced in multi-temporal satellite imagery, which contains rich semantic transitions (Figure~\ref{fig:definition}). Although explicit labels are unavailable during pre-training, these historical transitions provide semantic supervision. Therefore, we propose \textbf{Temporal Semantic Guidance}, which encourages the distance between temporal representations $\mathbf{z}_{t}$ and $\mathbf{z}_{t+\delta t}$ to reflect both the \textit{extent} and \textit{type} of actual urban changes. This constraint organizes the embedding space along meaningful semantic directions, enabling interpretable, high-level geo-semantic representations.

\subsubsection{Cumulative Semantic Signal Construction}

Before pre-training, we derive a semantic change metric in three steps: (1) extract visual entities with vision foundation models, (2) accumulate intermediate urban transitions, and (3) project these transitions into a semantic latent space. Pseudocode is provided in Appendix~\ref{temporal_guidance}. 

\textit{Step 1: Yearly change extraction}. For a geospatial tile observed in year $t$, we first generate a semantic map $\mathcal{S}_{t} \in \mathbb{R}^{H \times W}$ over land-use classes $\mathcal{C}$ (e.g., vegetation, water, buildings) using Grounded-SAM \cite{ren2024grounded}. Based on the detected change regions, we compute a yearly semantic change vector $\boldsymbol{\rho}^{(t, t+1)} \in \mathbb{R}^{|\mathcal{C}|}$ that quantifies the proportion of change for each semantic category. For each category $c \in \mathcal{C}$, the change ratio is obtained by measuring the overlap between the semantic map and the binary change mask, and normalizing by the total pixel area $|\Omega|$, as shown in Eq \ref{yearly_ratio}:
\begin{equation} 
{\rho}_{c}^{(t, t+1)} = \frac{1}{|\Omega|} \sum_{p \in \Omega} \left( \operatorname{Selector}_c \left( \mathcal{S}_{t}(p)\right) \cdot \mathcal{M}^{(t, t+1)}(p) \right), 
\label{yearly_ratio}
\end{equation}
where $\Omega$ denotes the pixel space, $p$ denotes the pixel, and $\operatorname{Selector}_c (\cdot)$ denotes the selected mask for category $c$.

\begin{figure}[!t]
    \centering
    \captionsetup{font={small}}
    \includegraphics[width=\columnwidth]{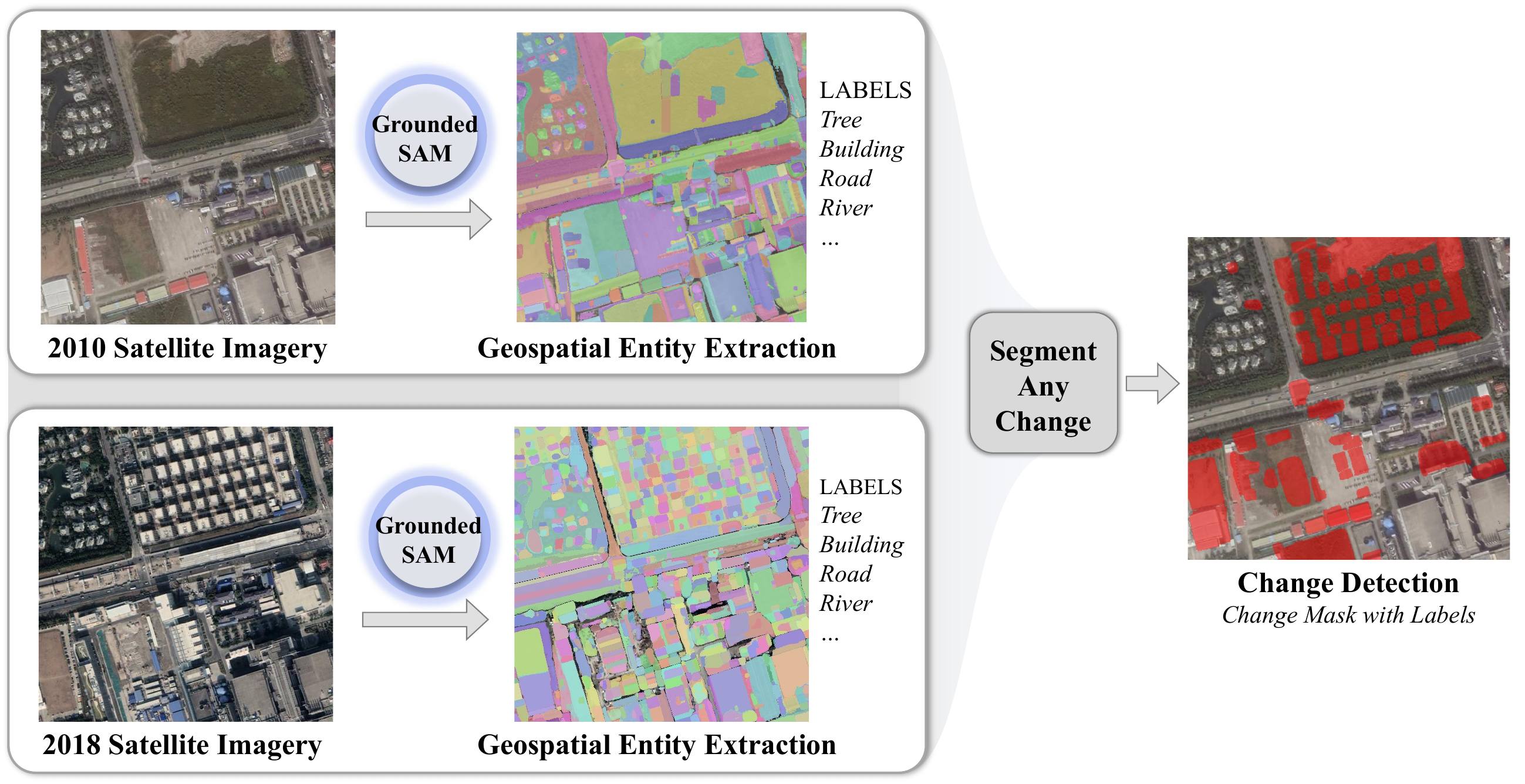}
    \caption{Joint modeling spatial and temporal dimensions from the perspectives of spatial proximity and temporal heterogeneity.}
    \label{fig:Temporal_changes}
    
\end{figure}

\textit{Step 2: Semantic Accumulation}. Urban region change is a continuous and cumulative process, where intermediate states contain crucial information (e.g., \textit{barren} $\to$ \textit{construction} $\to$ \textit{building}), while direct comparison between two time points misses these nuances and results in high computational complexity $\mathcal{O}(T^2)$. To capture the total semantic changes over an arbitrary time interval $[t, t+\delta t]$, we aggregate the yearly change vectors $\boldsymbol{\rho}^{(t, t+1)}$ to model intermediate transitions, with linear complexity $\mathcal{O}(T)$, obtaining the cumulative change vector $\boldsymbol{\rho}^{(t, t+\delta t)}$ , as shown in Eq. \ref{aggregate_ratio}: 
\begin{equation} 
\boldsymbol{\rho}^{(t, t+\delta t)} = \sum_{k=t}^{t+\delta t - 1} \boldsymbol{\rho}^{(k, k+1)}, 
\label{aggregate_ratio}
\end{equation}

\textit{Step 3: Semantic Projection.} To bridge the gap between low-level pixels and high-level semantics, we project this cumulative change vector into a semantic latent space to represent the accumulative semantic change magnitude. Let $\mathbf{C} \in \mathbb{R}^{|\mathcal{C}| \times d}$ denote the class embeddings generated by a frozen text encoder \cite{bert}. We compute the change embedding $\mathbf{e}^{(t, t+\delta t)}$ by performing a weighted aggregation of class embeddings based on their change ratios:

\begin{equation}
    \mathbf{e}^{(t, t+\delta t)} = \boldsymbol{\rho}^{(t, t+\delta t)} \cdot  \mathbf{C} \in \mathbb{R}^{d}.
    \label{ratio_emb}
\end{equation}

\subsubsection{Dual-Objective Temporal Optimization}

To make embedding distances reflect the direction and magnitude of real-world semantic changes, we use a dual-objective strategy. It aligns feature similarities with quantitative change signals while preserving instance discrimination for stable training.

\textit{1) Temporal Semantic Alignment}. To enhance interpretability, we constrain  the latent distance between two temporal views $\mathbf{z}_{t}$ and $\mathbf{z}_{t+\delta t}$ to be aligned with the cumulative semantic change magnitude $\|\mathbf{e}^{(t, t+\delta t)}\|_2$. Therefore, we define a soft label $y_{i, j} \in (0, 1)$ for each temporal pair based on the norm of change embeddings, using a hyperbolic tangent function to map change embeddings into a similarity space:  
\begin{equation}
    y_{i, j} = 1 - \tanh(\|\mathbf{e}^{(t, t+\delta t)}\|_2). 
    \label{change_norm}
\end{equation}

The soft label $y_{i,j}$ defines the target similarity: $y_{i,j} \approx 1$ indicates temporal stability and pulls the representations closer, whereas $y_{i,j} \to 0$ indicates substantial change and pushes them apart. We align visual similarity with these labels using the binary cross-entropy objective in Eq.~\ref{T_1}. Consequently, latent distances reflect the magnitude of real-world semantic changes.
\begin{multline}
    \mathcal{L}_{T_1} = - \sum_{(i,j) \in \mathcal{P}} \bigl[ y_{i,j} \log \sigma(\mathbf{z}_i \cdot \mathbf{z}_j / \tau) \\
    + (1 - y_{i,j}) \log (1 - \sigma(\mathbf{z}_i \cdot \mathbf{z}_j / \tau)) \bigr].
\label{T_1}
\end{multline}

\textit{2) Temporal Instance Discrimination}. To ensure robust learning, representations should maintain instance-level discriminability and stability. Empirical evidence indicates that training solely with a BCE-style objective leads to instability and slow convergence \cite{supcon}, and this issue is exacerbated by noise in the foundation model-derived embeddings $\mathbf{e}$. Thus, we introduce the InfoNCE loss \cite{mocov3} to leverage stable and discriminative gradients, thereby ensuring reliable convergence. In this contrastive task, for an anchor $\mathbf{z}_{i}$, the positive sample $\mathbf{z}_{i}^{+}$ is selected via augmentation to enforce representation invariance. Crucially, we introduce a set of temporal negatives $\mathcal{Z}_i^{-} = \left\{ \mathbf{z}_{i, t'}^- \mid \|\mathbf{e}^{(t, t')}\|_2 \neq 0 \right\}$, representing the same location at different time steps but exhibiting distinct semantic changes. By contrasting against these hard temporal negatives, the model is compelled to recognize significant urban changes. This objective is minimized by: 
\begin{equation}
    \mathcal{L}_{T_2} = -\sum_{i=1}^{\mathcal{B}} \log \frac{
        \exp \left( \langle \mathbf{z}_i, \mathbf{z}_{i}^{+} \rangle  \right)
    }{
        \exp \left( \langle \mathbf{z}_i, \mathbf{z}_{i}^{+} \rangle  \right) + 
        \sum_{\mathcal{Z}^{-}_i} \exp \left( \langle \mathbf{z}_i, \mathbf{z}_i^- \rangle \right)
    }.
    \label{eq:l_t2}
\end{equation}
The integration of \textit{Temporal Semantic Alignment} and \textit{Temporal Instance Discrimination} establishes a structurally robust representation space, enabling the model to learn discriminative features with faithful semantics. 
  
\subsection{Spatial-Temporal Alignment}

While \textit{Spatial Neighborhood Modeling} enhances generalization by capturing spatial correlations and \textit{Temporal Semantic Guidance} improves interpretability by characterizing semantic shifts, optimizing them independently may introduce \textit{false positives} in the embedding space. In real-world urban systems, visually similar regions may possess distinct latent semantics, which are only revealed through long-term urban development rather than static observation. An inherent dependency therefore exists: spatial similarity should be conditioned on whether regions exhibit consistent semantic changes over time, while modeling spatial and temporal dimensions in isolation overlooks this dependency. To resolve this, we propose the \textit{Spatial-Temporal Alignment} as a corrective mechanism that refines spatial similarity using temporal semantic changes.

Specifically, for a given anchor $I_i$, we identify its most similar neighbor $I_j \in \mathcal{N}_i$ based on cosine similarity, i.e., $j = \arg\max_{j \in \mathcal{N}_i} \mathbf{z}_i \cdot \mathbf{z}_j$. We then identify the timestamp $t'$ with the maximal semantic divergence from the current state and construct normalized semantic change vectors $\boldsymbol{\Delta}_i = \operatorname{norm}(\mathbf{z}_i - \mathbf{z}^{t'}_i)$ and $\boldsymbol{\Delta}_j = \operatorname{norm}(\mathbf{z}_j - \mathbf{z}^{t'}_j)$. Spatial similarity is penalized when these change directions are misaligned, thereby encouraging discrimination between visually similar but semantically divergent regions (Eq.~\ref{eq:st_loss}).

\begin{equation} 
\mathcal{L}_{st} = \frac{1}{|\mathcal{B}|} \sum_{i \in \mathcal{B}} (1 - \langle \boldsymbol{\Delta}_i, \boldsymbol{\Delta}_j \rangle) \cdot \langle \mathbf{z}_i, \mathbf{z}_j \rangle. 
\label{eq:st_loss} 
\end{equation}

\subsection{Pre-training \& Fine-Tuning}
\textbf{Pre-training Stage}. CoST adopts a contrastive learning framework for pre-training, initialized from scratch. The encoder $\mathcal{E}$ is optimized via a weighted multi-objective loss: 

\begin{equation}
    \mathcal{L}_{\text {pretrain}}=\alpha \mathcal{L}_{\mathrm{S}}+\beta_1  \mathcal{L}_{\mathrm{T_1}}+\beta_2 \mathcal{L}_{\mathrm{T_2}}+\gamma \mathcal{L}_{\mathrm{ST}},
    \label{final_loss}
\end{equation}
where the coefficients $\alpha$, $\beta_1$, $\beta_2$, and $\gamma$ are tuned via grid search to balance the training objectives. Jointly modeling spatial and temporal dynamics yields generalizable and interpretable representations. 

\noindent
\textbf{Fine-Tuning Stage}.
\label{sec:finetune_setting}
We adopt a linear probing strategy where the pre-trained encoder remains frozen. We attach the lightweight task-specific heads to evaluate the representations across three downstream tasks: 
\begin{itemize}[leftmargin=*]
    \item \textbf{Static Indicator Prediction}. For socio-economic indicator prediction from a single target year, the target representation $\mathbf{z}_T$ is fed into an MLP regression head:
    \begin{equation}
        \hat{y}=\operatorname{MLP}_{\mathrm{reg}}(\mathbf{z}_T).
    \end{equation}

    \item \textbf{Dynamic Indicator Prediction}. To incorporate historical urban dynamics, we further use the temporal representation sequence $\mathbf{Z}=\{\mathbf{z}_1,\dots,\mathbf{z}_{T-1}\}$ as auxiliary input. An LSTM encoder summarizes the historical sequence, and its output is combined with the target-year representation to predict the indicator:
    \begin{equation}
        \hat{y}=\operatorname{MLP}_{\mathrm{base}}(\mathbf{z}_T)+\operatorname{MLP}_{\mathrm{res}}\bigl(\operatorname{LSTM}(\mathbf{Z})\bigr).
    \end{equation}

    \item \textbf{Land-Use Classification}. For scene-level land-use recognition, the representation $\mathbf{z}$ is fed into a lightweight MLP classifier followed by a softmax layer, as shown in Eq.~\ref{eq:softmax}. The classifier is optimized using cross-entropy loss, evaluating whether the learned representation preserves discriminative semantic cues for region categorization.
    \begin{equation}
        \hat{y}=\operatorname{Softmax}(\operatorname{MLP}_{\mathrm{cls}}(\mathbf{z})).
        \label{eq:softmax}
    \end{equation}

    \item \textbf{Change Detection}. For bi-temporal change detection, we extract representations from two timestamps and concatenate them as $[\mathbf{z}_t;\mathbf{z}_{t'}]$. The fused feature is then passed to a lightweight U-Net decoder to predict the binary change map, as shown in Eq.~\ref{eq:change_mlp}. The decoder is trained with binary cross-entropy loss to assess whether the learned representations retain fine-grained temporal change information.
    \begin{equation}
        \hat{y}=\mathcal{U}([\mathbf{z}_t;\mathbf{z}_{t'}]).
        \label{eq:change_mlp}
    \end{equation}
\end{itemize}

\section{Experiments}
\begin{figure}[!t]
    \centering
    \captionsetup{font={small}}
    \includegraphics[width=\columnwidth]{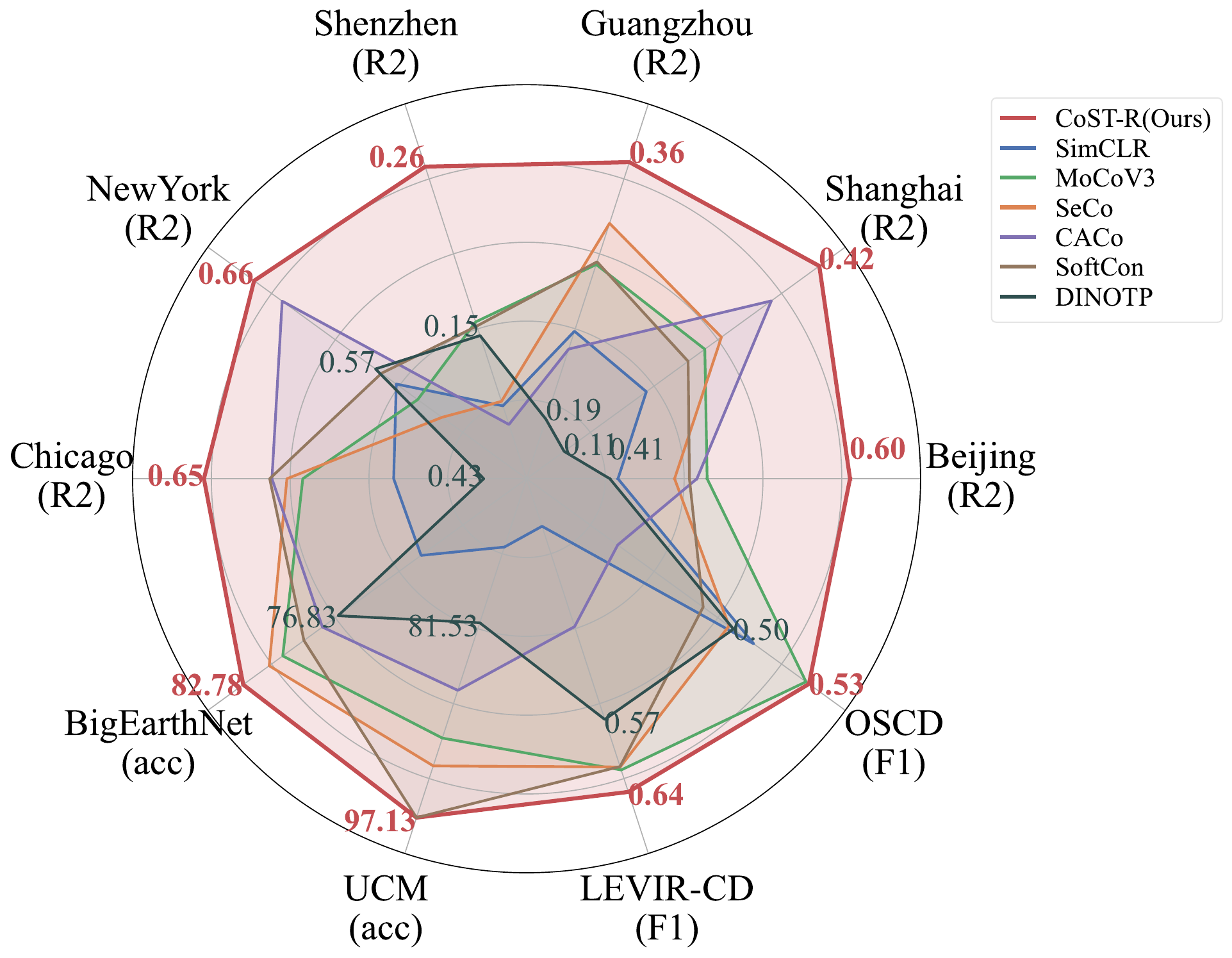}
    \caption{Overall performance of CoST-R across various downstream tasks.}
    \label{fig:radar}
\end{figure}

In this section, we evaluate the proposed CoST to investigate the following research questions (\textbf{RQs}):

\begin{itemize}[leftmargin=*]
    \item \textbf{RQ1:} Does CoST outperform state-of-the-art self-supervised methods and demonstrate \textit{generalizability} across diverse tasks and real-world applications in complex urban environments?

    \item \textbf{RQ2:} Do the learned representations possess semantic \textit{interpretability}, where the latent space effectively captures and aligns with meaningful spatial-temporal transitions?

    \item \textbf{RQ3:} What are the contributions of individual components to the overall performance of the CoST?


    \item \textbf{RQ4:} How do temporal dynamics regularize semantic consistency and improve domain invariance across cities?

\end{itemize}

\begin{table}[htbp]
\centering
\captionsetup{font=normalfont}
\setlength{\abovecaptionskip}{2pt}
\setlength{\belowcaptionskip}{2pt} 
\renewcommand{\arraystretch}{1}
\caption{Dataset statistics. The first five datasets are used for pre-training, and the \textit{Chicago} dataset is held out from pre-training for a fair generalization test.}
\label{tab:dataset_statistics}

\resizebox{\linewidth}{!}{ 
    \begin{tabular}{c|cc|c}
    \toprule
    \multirow{2}{*}{\textbf{Dataset}} & \multicolumn{2}{c|}{\textbf{Coverage in Geospace}} & \multirow{2}{*}{\textbf{Satellite}} \\
    \cmidrule{2-3}
    & \textit{Bottom-left} & \textit{Top-right} & \textbf{Image} \\
    \midrule
    Beijing & 39.91°N, 116.20°E & 40.24°N, 116.53°E & 4,218 $\times$ 11 \\
    Guangzhou & 22.78°N, 113.15°E & 23.36°N, 113.51°E & 3,726 $\times$ 11 \\
    Shanghai & 30.87°N, 121.24°E & 31.40°N, 121.68°E & 3848 $\times$ 11 \\
    Shenzhen & 22.52°N, 113.84°E & 22.67°N, 114.25°E & 2,772 $\times$ 11 \\
    New York & 40.52°N, 73.95°W & 40.80°N, 73.79°W & 1,728 $\times$ 11 \\
    Chicago & 41.84°N, 87.74°W & 41.96°N, 87.57°W & 754 $\times$ 11 \\
    \bottomrule
    \end{tabular}
} 
\end{table}

\noindent
\textbf{Pre-training Dataset.}
We pre-train CoST on a spatial-temporal satellite imagery dataset collected from Google Earth, covering five major cities: \textit{Beijing}, \textit{Shanghai}, \textit{Guangzhou}, \textit{Shenzhen}, and \textit{New York}, over the period from 2010 to 2020. Each sample corresponds to a geospatial region with aligned multi-year observations, which enables the model to jointly learn spatial structure and temporal dynamics. Dataset statistics are summarized in Tab.~\ref{tab:dataset_statistics}. To evaluate cross-city generalization, we further construct a held-out \textit{Chicago} dataset, which is excluded from pre-training and used only for downstream evaluation.

\noindent
\textbf{Downstream Datasets.}
We evaluate CoST on three types of downstream tasks to assess its effectiveness across socio-economic inference, semantic classification, and temporal change understanding. 
\begin{itemize}[leftmargin=*]
    \item \textbf{Socio-economic Indicator Prediction.} We evaluate urban indicator prediction using two representative socio-economic signals: population density data from WorldPop\footnote{https://hub.worldpop.org/} and GDP data from a publicly available global dataset\footnote{https://www.nature.com/articles/s41597-022-01322-5}. As both datasets provide multi-year observations, they support evaluations under both static prediction and dynamic prediction settings using historical representation sequences. This task assesses whether the learned representations capture cross-sectional socio-economic semantics at a given time point and their temporal dynamics over time. 
    
    \item \textbf{Land-Use Classification.} We adopt two benchmark datasets for semantic scene understanding: UC Merced Land-Use~\cite{ucm}, which contains aerial scene categories for single-label classification, and BigEarthNet~\cite{sumbul2019bigearthnet}, a large-scale multi-label remote sensing benchmark with diverse land-cover types. This setting evaluates whether the learned embeddings retain discriminative semantics for region categorization.
    
    \item \textbf{Change Detection.} We evaluate temporal sensitivity on two widely used change detection datasets, OSCD~\cite{oscd} and LEVIR-CD~\cite{levircd}. Both datasets provide bi-temporal observations with pixel-level supervision, enabling us to assess whether the learned representations encode fine-grained change information for downstream segmentation.
\end{itemize}

\noindent
\textbf{Baselines.}
We compare CoST with self-supervised and geospatial representation learning baselines using both CNN and ViT encoders. These include general visual pre-training methods (SimCLR~\cite{simclr}, MoCoV3~\cite{mocov3}, DINOv2~\cite{oquab2023dinov2}, and SoftCon~\cite{wang2024softcon}), remote-sensing methods (SeCo~\cite{manas2021seco}, CACo~\cite{mall2023caco}, READ~\cite{read}, PG-SimCLR~\cite{pgsimclr}, SatMAE~\cite{cong2022satmae}, and ScaleMAE~\cite{reed2023scalemae}), and temporal transformer variants (DiNOTP~\cite{wanyan2024dinomc}, DiNOTP-ViT, and SoftCon-ViT). Appendix~\ref{baselines} provides brief descriptions.

\noindent
\textbf{Metrics.}
We adopt task-specific evaluation metrics following standard practice. For socio-economic indicator prediction, we report $R^2$ and mean squared error (MSE) to measure regression accuracy. For land-use classification, we report Top-1 accuracy. For change detection, we report the F1 score to evaluate the quality of predicted change regions.

\noindent
\textbf{Implementation Details.}
We implement two variants of CoST: \textbf{CoST-R}, based on a ResNet50 backbone, and \textbf{CoST-V}, based on a ViT-B/16 backbone. Both variants use a feature dimension of $d=128$ and follow the same pre-training and downstream evaluation protocol unless otherwise specified. All models are pre-trained for 3,000 epochs on a server equipped with two NVIDIA RTX A6000 GPUs. Detailed optimization settings, loss hyperparameters, and downstream fine-tuning configurations are provided in Appendix~\ref{pretraining_details} and Appendix~\ref{finetuning_details}.

\subsection{Overall Performance (RQ1)}

\begin{table*}[htbp]
\centering
\captionsetup{font=small}
\caption{Performance on urban indicator prediction (2020). \textbf{Bold} and \underline{underline} denote the best and second-best results, respectively. Both variants of our model (CoST-R and CoST-V) are highlighted in \textbf{bold} when they achieve the top two positions.}
\label{tab:indicator_tab}

\renewcommand{\arraystretch}{1.3}
\setlength{\tabcolsep}{5.5pt}

\resizebox{\textwidth}{!}{

\begin{tabular}{l cccc cccc cccc cccc}
\toprule

\multirow{3.5}{*}{\textbf{Model}} & 
\multicolumn{4}{c}{\textbf{Beijing}} & 
\multicolumn{4}{c}{\textbf{Guangzhou}} & 
\multicolumn{4}{c}{\textbf{New York}} & 
\multicolumn{4}{c}{\textit{\textbf{Chicago}}} \\

\cmidrule(lr){2-5} \cmidrule(lr){6-9} \cmidrule(lr){10-13} \cmidrule(lr){14-17}

 & \multicolumn{2}{c}{\textbf{GDP}} & \multicolumn{2}{c}{\textbf{Population}} & 
      \multicolumn{2}{c}{\textbf{GDP}} & \multicolumn{2}{c}{\textbf{Population}} & 
      \multicolumn{2}{c}{\textbf{GDP}} & \multicolumn{2}{c}{\textbf{Population}} & 
      \multicolumn{2}{c}{\textbf{GDP}} & \multicolumn{2}{c}{\textbf{Population}} \\

\cmidrule(lr){2-3} \cmidrule(lr){4-5} 
\cmidrule(lr){6-7} \cmidrule(lr){8-9} 
\cmidrule(lr){10-11} \cmidrule(lr){12-13} 
\cmidrule(lr){14-15} \cmidrule(lr){16-17}

 & \textit{R$^2$} & \textit{MSE}$\downarrow$ & \textit{R$^2$} & \textit{MSE}$\downarrow$ & 
      \textit{R$^2$} & \textit{MSE}$\downarrow$ & \textit{R$^2$} & \textit{MSE}$\downarrow$ & 
      \textit{R$^2$} & \textit{MSE}$\downarrow$ & \textit{R$^2$} & \textit{MSE}$\downarrow$ & 
      \textit{R$^2$} & \textit{MSE}$\downarrow$ & \textit{R$^2$} & \textit{MSE}$\downarrow$ \\
\midrule

\rowcolor{gray!10}
SimCLR & 0.420 & 0.584 & 0.412 & 0.570 & 0.248 & 0.684 & 0.163 & 0.467 & 0.552 & 0.363 & 0.608 & 0.353 & 0.501 & 0.561 & 0.467 & 0.522 \\

MoCoV3 & {0.488} & {0.460} & {0.504} & 0.510 & 0.293 & 0.781 & 0.390 & \underline{0.447} & 0.535 & 0.434 & 0.594 & 0.391 & 0.570 & 0.367 & 0.597 & 0.447 \\

\rowcolor{gray!10}
DINOv2 & 0.346 & 0.605 & 0.456 & 0.544 & 0.151 & 0.845 & 0.296 & 0.771 & 0.593 & 0.435 & 0.622 & {0.327} & 0.487 & 0.499 & 0.661 & 0.357 \\

SeCo & 0.463 & 0.551 & 0.462 & 0.497 & {0.320} & 0.728 & \underline{0.472} & 0.491 & 0.516 & 0.502 & 0.606 & 0.375 & 0.583 & {0.342} & 0.531 & 0.415 \\

\rowcolor{gray!10}
CACo & 0.480 & 0.543 & 0.477 & {0.489} & 0.237 & 0.714 & 0.229 & 0.600 & \underline{0.642} & 0.413 & \underline{0.666} & 0.361 & {0.594} & 0.431 & \underline{0.669} & 0.291 \\

SoftCon & 0.474 & \underline{0.424} & 0.472 & 0.507 & 0.295 & 0.862 & 0.267 & 0.837 & 0.564 & 0.382 & 0.624 & 0.379 & 0.596 & 0.438 & 0.601 & \underline{0.256} \\

\rowcolor{gray!10}
DiNOTP & 0.413 & 0.585 & 0.364 & 0.706 & 0.190 & 0.774 & 0.211 & 1.215 & 0.569 & 0.394 & 0.543 & 0.408 & 0.433 & 0.727 & 0.463 & 0.640 \\

READ & 0.217 & 0.805 & 0.182 & 0.805 & 0.291 & 0.679 & 0.327 & 0.610 & - & - & - & - & - & - & - & - \\

\rowcolor{gray!10}
PG-SimCLR & 0.418 & 0.572 & 0.468 & \underline{0.439} & 0.266 & \underline{0.616} & - & - & - & - & - & - & - & - & - & - \\

SatMAE & 0.489 & 0.593 & \underline{0.513} & 0.465 & \underline{0.339} & 1.005 & 0.382 & 0.872 & 0.639 & \textbf{0.320} & 0.612 & 0.371 & \underline{0.607} & \underline{0.297} & 0.580 & 0.360 \\

\rowcolor{gray!10}
ScaleMAE & \underline{0.498} & 0.479 & 0.511 & 0.501 & 0.208 & 0.941 & 0.333 & 0.771 & 0.639 & 0.389 & 0.629 & \underline{0.319} & 0.588 & 0.399 & 0.628 & 0.383 \\

\rowcolor{green!10}
{CoST-R} & \textbf{0.596} & \textbf{0.393} & \textbf{0.565} & \textbf{0.410} & \textbf{0.361} & \textbf{0.498} & \textbf{0.553} & \textbf{0.429} & \textbf{0.664} & \underline{0.339} & \textbf{0.698} & \textbf{0.273} & \textbf{0.646} & \textbf{0.276} & \textbf{0.691} & \textbf{0.253} \\

\rowcolor{green!10}
{CoST-V} & \textbf{0.561} & 0.477 & \textbf{0.545} & \textbf{0.430} & \textbf{0.378} & 0.738 & {0.409} & {0.460} & {0.630} & {0.370} & \textbf{0.678} & {0.322} & \textbf{0.642} & 0.331 & {0.655} & {0.355} \\

\bottomrule
\end{tabular}
}
\end{table*}

To evaluate the effectiveness and generalizability of CoST, we conduct extensive experiments against baselines across diverse tasks. Fig \ref{fig:radar} presents a holistic overview of CoST-R's performance in three distinct domains. Detailed quantitative comparisons are provided in Tab.~\ref{tab:indicator_tab} and Tab.~\ref{tab:cls_cd_comparison}. 

\noindent
\textbf{Urban Indicator Prediction.} CoST (CoST-R \& CoST-V) achieves state-of-the-art performance in urban indicator prediction across multiple cities. In New York City, CoST-R attains an $R^2$ of 0.664 for GDP and 0.698 for population, outperforming prior methods by over 5\%. Similar improvements are observed in Shanghai and Shenzhen (see Appendix~\ref{sec:experiment_results}). To evaluate cross-region generalization under domain shift, we further test on \textit{Chicago}, which is excluded from pre-training. As shown in Tab.~\ref{tab:indicator_tab}, CoST achieves strong transfer results, with GDP and population $R^2$ of 0.646 and 0.691, respectively, surpassing region-dependent baselines and demonstrating robust generalization across urban environments.

\begin{table}[h]
\captionsetup{font=small}
\centering
\renewcommand{\arraystretch}{1.15} 
\setlength{\tabcolsep}{12pt} 

\caption{Performance of different models in classification and change detection tasks. The best results are \textbf{bold}, and the second-best results are \underline{underlined}.}
\label{tab:cls_cd_comparison}

\resizebox{\columnwidth}{!}{%
\begin{tabular}{lcccc}
\toprule

\multirow{3.5}{*}{\textbf{Methods}} & \multicolumn{2}{c}{\textbf{Classification}} & \multicolumn{2}{c}{\textbf{Change Detection}} \\

\cmidrule(lr){2-3} \cmidrule(lr){4-5}

& \textbf{UCM} & \textbf{BigEarthNet} & \textbf{OSCD} & \textbf{LEVIR-CD} \\
& \textit{top-1 Acc} &\textit{mAP} & \textit{F1 Score} & \textit{F1 Score} \\

\midrule

\rowcolor{gray!10}
SimCLR  & 75.47 & 71.62 & 0.5069 & 0.3944 \\

MoCoV3  & 90.76 & 80.30 & \underline{0.5315} & \underline{0.6220} \\

\rowcolor{gray!10}
SeCo    & 92.99 & 81.15 & 0.4953 & 0.6192 \\

CACo    & 86.94 & 77.81 & 0.4429 & 0.4881 \\

\rowcolor{gray!10}
SoftCon & {97.13} & 78.97 & 0.4831 & 0.6188 \\

DiNOTP  & 81.53 & 76.83 & 0.4974 & 0.5748 \\

\rowcolor{gray!10}
SatMAE  & 92.36 & 76.07 & -- & -- \\

ScaleMAE& \underline{98.41} & \underline{82.57} & -- & -- \\

\rowcolor{green!10}
CoST-R    & {97.13} & \textbf{82.78} & \textbf{0.5329} & \textbf{0.6425} \\

\rowcolor{green!10}
CoST-V & \textbf{98.73} & \textbf{83.55} & -- & -- \\

\bottomrule
\end{tabular}%
}
\end{table}

\noindent
\textbf{Classification \& Change Detection.}
We evaluate the learned representations on land-use classification and change detection (Tab.~\ref{tab:cls_cd_comparison}). For classification, CoST-R and CoST-V achieve 97.13\% and 98.73\% top-1 accuracy on UCM, respectively, while remaining competitive on the multi-label BigEarthNet dataset. For change detection, CoST-R surpasses specialized bi-temporal methods~\cite{manas2021seco,mall2023caco}, achieving an F1 score of 0.6425 on LEVIR-CD, demonstrating robust cross-task generalization of CoST.

\subsection{Representation Interpretability (RQ2)}

\begin{figure}[t]
    \centering
    \captionsetup{font={small}}
    \includegraphics[width=\columnwidth]{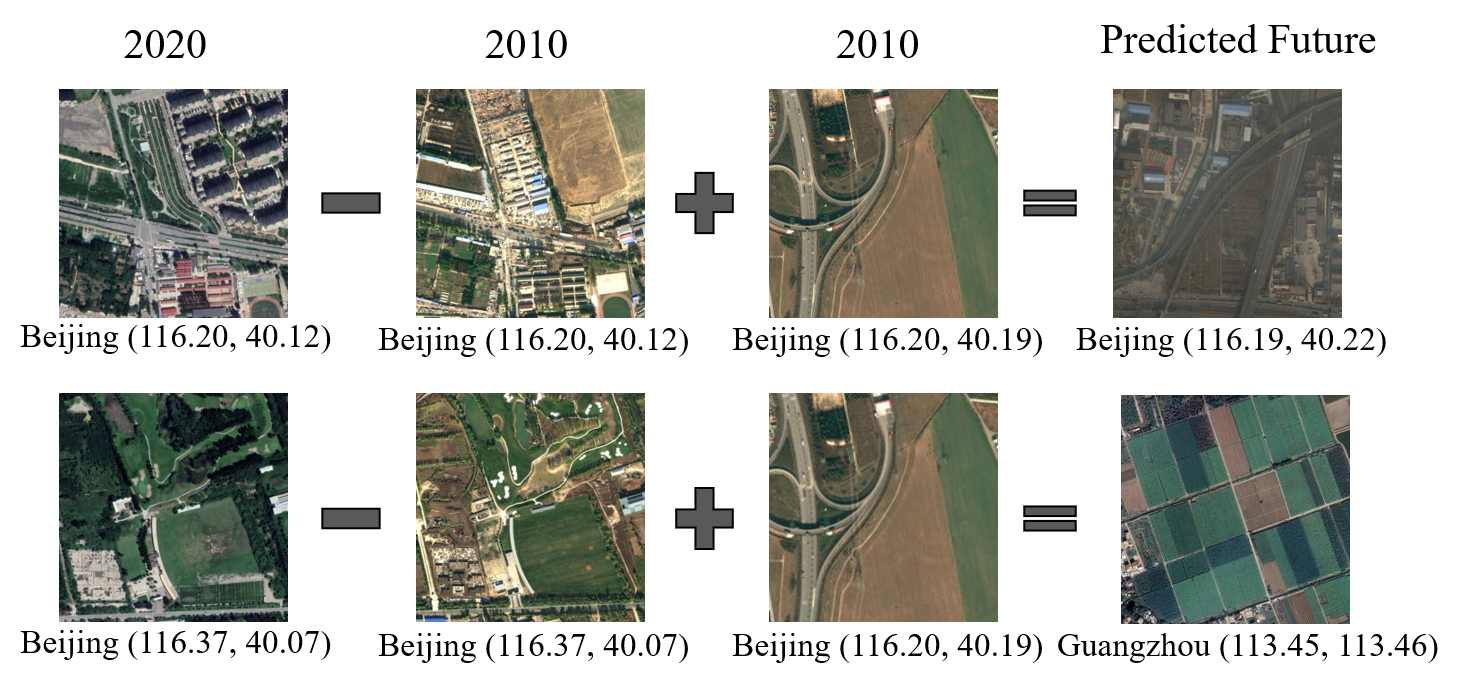}
    \caption{Visualization of embedding arithmetic. It shows that CoST captures high-level semantic concepts via vector operations. Predictable semantic transformations demonstrate that the learned latent space is disentangled and interpretable. }
    \label{fig:embedding_visualization}
\end{figure}

To evaluate the semantic interpretability of the learned representations, we investigate whether the latent space captures high-level concepts through vector arithmetic, as shown in Figure~\ref{fig:embedding_visualization}(a). If visual semantics are aligned with the embedding space, algebraic operations should induce predictable semantic changes. Specifically, subtracting one attribute (e.g., built-up areas) and adding another (e.g., farmland) retrieves images in which the target semantic components change while the surrounding context is preserved. This consistency indicates that CoST captures meaningful semantic directions rather than memorizing pixel-level patterns, demonstrating that the learned representations are semantically disentangled and interpretable.

\subsection{Ablation Study (RQ3)}

\begin{figure}[t]
    \centering
    \captionsetup{font={small}}
    \includegraphics[width=\columnwidth]{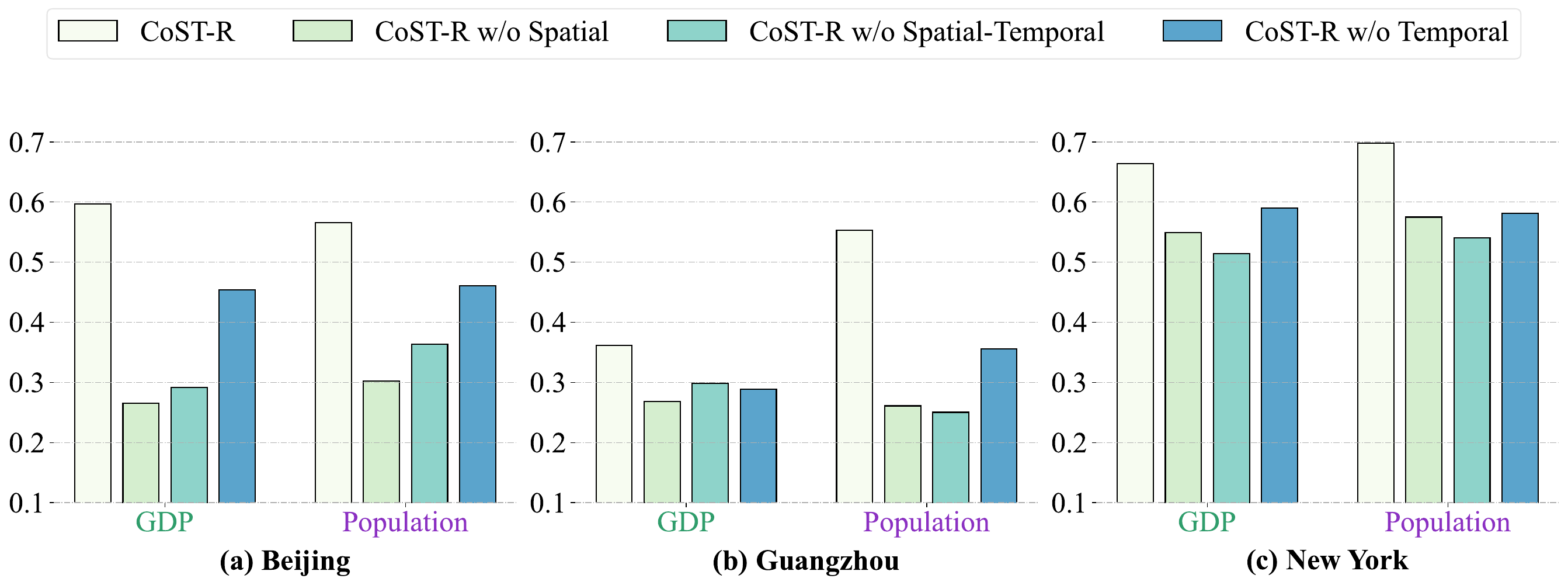}
    \caption{Ablation study of three CoST components: (a) \textit{w/o Spatial}, (b) \textit{w/o Temporal}, and (c) \textit{w/o ST}.}
    \label{fig:ablation_top}
\end{figure}

To evaluate the contribution of each component in CoST, we conduct an ablation study on socio-economic prediction tasks with three variants: (a) \textit{w/o Spatial}, (b) \textit{w/o Temporal}, and (c) \textit{w/o ST}, as shown in Fig.~\ref{fig:ablation_top}. Removing spatial neighborhood modeling (\textit{w/o Spatial}) leads to a pronounced performance drop, with $R^2$ decreasing by 15–30\% across cities, highlighting the importance of spatial aggregation in capturing shared urban contexts following Tobler’s first law \cite{tobler1970computer}. The performance decline in \textit{w/o Temporal} further demonstrates that temporal semantics provide critical supervision for representation learning. Finally, the performance of \textit{w/o ST} suggests that spatial–temporal alignment is essential for avoiding the aggregation of visually similar yet functionally distinct regions, enabling accurate indicator regression.

\subsection{Qualitative Analysis (RQ4)}

\begin{figure}[t]
    \centering
    \captionsetup{font={small}}
    \includegraphics[width=\columnwidth]{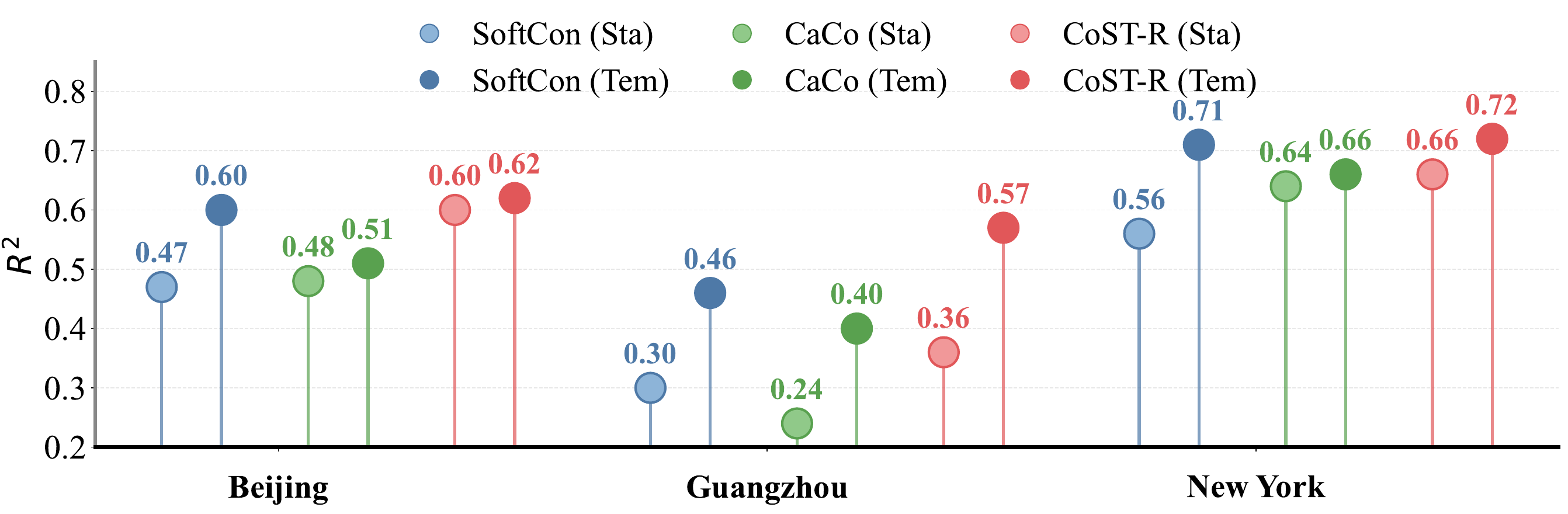}
    \caption{Urban indicator prediction using time-series satellite imagery in Beijing, Guangzhou, and New York City. \textit{Sta} and \textit{Temp} denote static and dynamic prediction, respectively.}
    \label{fig:temporal_prediction}
\end{figure}

\textit{Dynamic Urban Profiling}. To assess CoST’s ability to model long-term urban development, we perform dynamic urban profiling by leveraging historical representation sequences (2010–2019) to predict 2020 indicators. Detailed task descriptions and experimental settings are provided in Sec.~\ref{sec:finetune_setting} and Appendix~\ref{finetuning_details}. As shown in Figure~\ref{fig:temporal_prediction}, incorporating temporal dynamics (Tem) consistently outperforms static snapshots (Sta) in terms of $R^2$ across all cities. Notably, CoST-R achieves the best performance among temporal-aware baselines, reaching $R^2$ scores of 0.62 in Beijing, 0.57 in Guangzhou, and 0.72 in New York. These results highlight the importance of historical sequences for accurate urban indicator estimation, as temporal accumulation enables the model to capture socio-economic development rather than relying on a single visual snapshot, leading to more coherent and predictive representations over time.

\begin{figure}[t]
    \centering
    \captionsetup{font={small}}
    \includegraphics[width=\columnwidth]{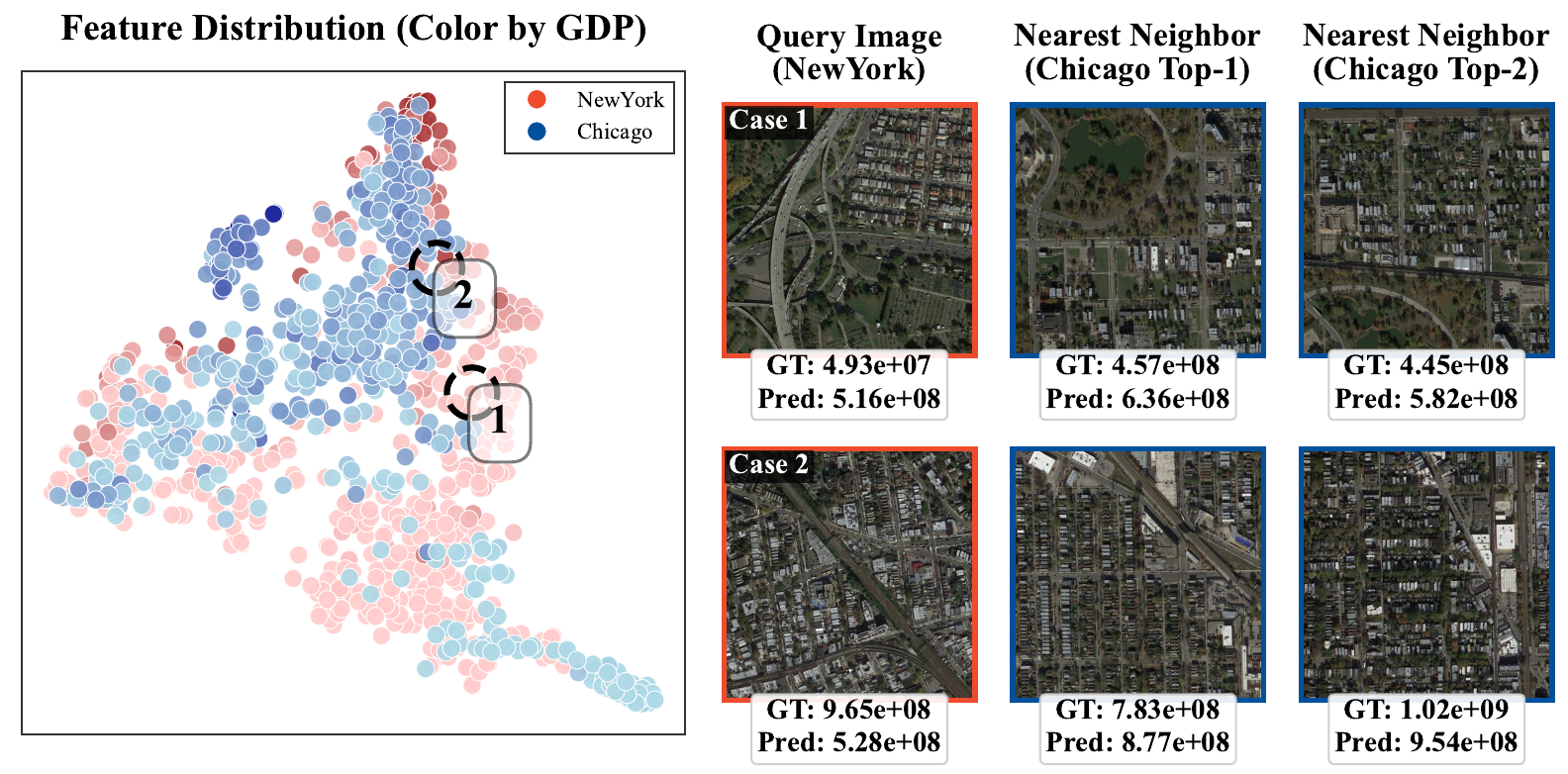}
    \caption{Domain adaptation of CoST between the source city and target city (Chicago). Color saturation represents GDP levels from {light (low GDP)} to {dark (high GDP)}. }
    \label{fig:qualitative_analysis_general}
\end{figure}

\noindent
\textit{Universal Domain Shift}. 
To understand how CoST handles visual disparities across urban environments, we perform the case-based retrieval analysis in Figure~\ref{fig:qualitative_analysis_general}. Despite substantial domain shifts, the latent representations align by socio-economic level, and the resulting clusters reflect GDP levels even for unseen data. In both suburban and urban cases, New York queries retrieve semantically consistent Chicago regions with matching ground-truth GDP profiles. This alignment indicates that CoST captures geographic structures that transfer to unseen regions.

\section{Conclusion and Future Work}

In this work, we have presented {CoST}, a unified framework designed to learn generalizable and interpretable geospatial representations from satellite imagery, effectively bridging the gap between visual features and high-level geo-semantics. Extensive experiments demonstrate that CoST achieves state-of-the-art performance across diverse downstream tasks. Building upon the continuous multi-year imagery used in this work, future research could further explore predictive representations for urban change prediction and image generation tasks, enabling more proactive urban planning and scenario simulation.

\begin{acks}
    This work is supported by the Guangdong Basic and Applied Basic Research Foundation (No. 2025A1515011994), the National Natural Science Foundation of China (No. 62402414), Guangdong Provincial Project 2025D03J0014, Guangzhou Municipal Science and Technology Project (No. 2023A03J0011), the Guangzhou Industrial Information and Intelligent Key Laboratory Project (No. 2024A03J0628), and Guangdong Provincial Key Lab of Integrated Communication, Sensing and Computation for Ubiquitous Internet of Things (No. 2023B1212010007).
\end{acks}

\appendix

\section{Experiment Results}
\label{sec:experiment_results}
We provide additional urban indicator results on Shanghai and Shenzhen, complementing the main-paper evaluation. The results show that CoST remains effective across both cities and both indicators.

\begin{table}[htbp]
\centering
\small 
\captionsetup{font=small}
\caption{Performance Comparison on \textbf{Shanghai} Dataset. \textbf{Bold} and \underline{underline} denote the best and second-best results, respectively. Both variants of our model (CoST-R and CoST-V) are highlighted in \textbf{bold} when they achieve the top two positions.}
\label{tab:shanghai} 

\setlength{\tabcolsep}{6pt} 
\renewcommand{\arraystretch}{1.2} 

\begin{tabular}{l|c|cc|cc}
\toprule
\multirow{2}{*}{\textbf{Model}} & \multirow{2}{*}{\textbf{Backbone}} & \multicolumn{2}{c|}{\textbf{GDP}} & \multicolumn{2}{c}{\textbf{Population}} \\
\cmidrule{3-6}
 & & \textbf{R$^2$} & \textbf{MSE}$\downarrow$ & \textbf{R$^2$} & \textbf{MSE}$\downarrow$ \\
\midrule
\rowcolor{gray!10}
SimCLR & R50 & 0.213 & 0.955 & 0.284 & 0.711 \\
MoCoV3 & R50 & 0.283 & \textbf{0.523} & \underline{0.379} & \textbf{0.443} \\
\rowcolor{gray!10}
DINOv2 & R50 & 0.179 & 0.692 & 0.293 & 0.736 \\
SeCo & R50 & 0.302 & \underline{0.542} & 0.298 & 0.919 \\
\rowcolor{gray!10}
CACo & R50 & \underline{0.362} & 0.860 & 0.305 & 0.734 \\
SoftCon & R50 & 0.263 & 0.857 & 0.332 & 0.685 \\
\rowcolor{gray!10}
DiNOTP & R50 & 0.115 & 0.984 & 0.222 & 0.679 \\
SatMAE & ViT-L & 0.404 & \textbf{0.515} & 0.338 & 0.712 \\
\rowcolor{gray!10}
ScaleMAE & ViT-B & 0.351 & 0.916 & 0.311 & 0.607 \\
\rowcolor{green!10}
CoST-R & R50 & \textbf{0.419} & 0.675 & \textbf{0.434} & 0.550 \\
\rowcolor{green!10}
CoST-V & ViT-B & \textbf{0.442} & 0.723 & {0.372} & \underline{0.505} 
\\ \bottomrule
\end{tabular}
\end{table}

\begin{table}[htbp]
\centering
\small 
\captionsetup{font=small}
\caption{Performance Comparison on \textbf{Shenzhen} Dataset. \textbf{Bold} and \underline{underline} denote the best and second-best results, respectively. Both variants of our model (CoST-R and CoST-V) are highlighted in \textbf{bold} when they achieve the top two positions.}
\label{tab:shenzhen}

\setlength{\tabcolsep}{6pt}
\renewcommand{\arraystretch}{1.2}
\begin{tabular}{l|c|cc|cc}
\toprule
\multirow{2}{*}{\textbf{Model}} & \multirow{2}{*}{\textbf{Backbone}} & \multicolumn{2}{c|}{\textbf{GDP}} & \multicolumn{2}{c}{\textbf{Population}} \\
\cmidrule{3-6}
 & & \textbf{R$^2$} & \textbf{MSE}$\downarrow$ & \textbf{R$^2$} & \textbf{MSE}$\downarrow$ \\
\midrule
\rowcolor{gray!10}
SimCLR & R50 & 0.099 & 0.832 & 0.296 & 0.623 \\
MoCoV3 & R50 & \underline{0.155} & 0.998 & \underline{0.382} & \underline{0.569} \\
\rowcolor{gray!10}
DINOv2 & R50 & 0.074 & 0.943 & 0.316 & 0.709 \\
SeCo & R50 & 0.102 & 0.894 & 0.274 & 0.653 \\
\rowcolor{gray!10}
CACo & R50 & 0.086 & 0.959 & 0.291 & 0.578 \\
SoftCon & R50 & 0.152 & 0.951 & 0.352 & 0.668 \\
\rowcolor{gray!10}
DiNOTP & R50 & 0.146 & \underline{0.791} & 0.320 & 0.718 \\
READ & R18 & 0.129 & 0.889 & 0.284 & 0.972 \\
\rowcolor{gray!10}
PG-SimCLR & R18 & 0.123 & 1.058 & - & - \\
SatMAE & ViT-L & 0.176 & \textbf{0.623} & 0.325 & 0.773 \\
\rowcolor{gray!10}
ScaleMAE & ViT-B & 0.163 & 0.796 & 0.329 & 0.725 \\
\rowcolor{green!10}
CoST-R & R50 & \textbf{0.258} & 0.967 & \textbf{0.463} & \textbf{0.496} \\
\rowcolor{green!10}
CoST-V & ViT-B & \textbf{0.273} & 0.856 & \textbf{0.359} & \textbf{0.591} 
\\ \bottomrule
\end{tabular}
\end{table}

\begin{figure}[h]
    \centering
    \includegraphics[width=\linewidth]{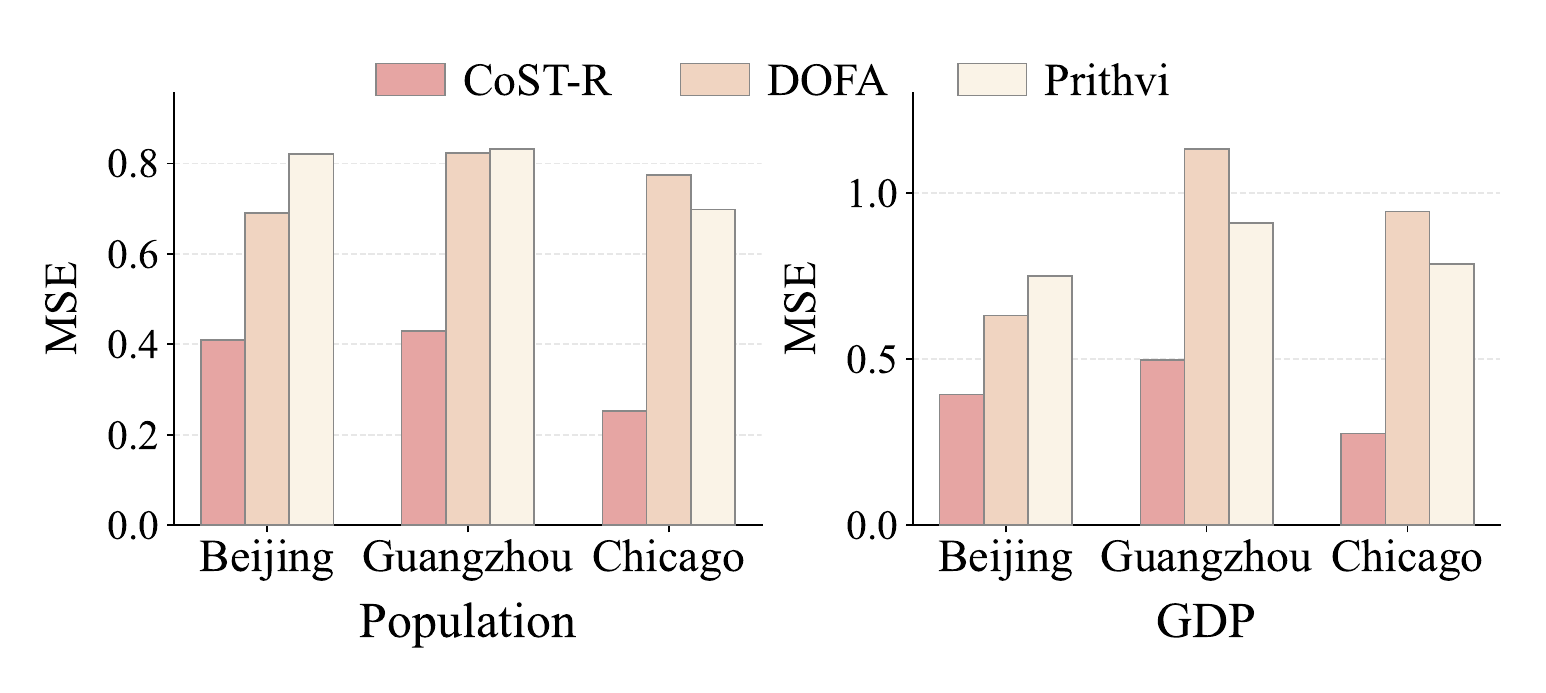}
    \caption{Performance comparison with RS foundation models across indicator predictions. CoST-R, Prithvi-100M, and DOFA are compared using GDP MSE and population MSE.}
    \label{fig:fm_comparison}
\end{figure}

We provide additional urban indicator results on the Shanghai and Shenzhen datasets, further supporting the claims in Section~4.2 and demonstrating CoST's robust profiling capability across diverse urban contexts. As shown in Table~\ref{tab:shanghai}, CoST-V achieves a GDP $R^2$ of 0.442 in Shanghai, improving over SatMAE (0.404), while CoST-R leads population prediction with an $R^2$ of 0.434. In Shenzhen, CoST-V obtains the highest GDP $R^2$ of 0.273, and CoST-R achieves the best population performance with an $R^2$ of 0.463 and an MSE of 0.496, indicating balanced performance across both indicators and cities. We further compare CoST-R with the remote-sensing foundation models Prithvi-100M and DOFA. As shown in Fig.~\ref{fig:fm_comparison}, CoST-R consistently achieves lower GDP and population MSE than both models in Beijing, Guangzhou, and the unseen Chicago setting, despite using a substantially smaller ResNet50 backbone.

\subsection{Pre-training Details}
\label{pretraining_details}
A frozen BERT encoder generates embeddings for 11 urban land-use categories: \texttt{dense houses, trees, grass, river, barren land, sidewalks, farmland, tall buildings, soil, crop fields, roads}. We use Adam ($\beta=(0.9,0.999)$, weight decay $1\times10^{-4}$) with a target learning rate of $3\times10^{-4}$, linearly warmed up over 20 epochs and then held constant. We use 32 groups per GPU (64 groups in total), with four elements per group. We set $num_{neg}=6$, $\tau=0.07$, and $\epsilon=0.1$, and weight the total loss with $(\alpha,\beta_1,\beta_2,\gamma)=(1.0,0.2,0.2,0.3)$.

\subsection{Fine-tuning details}
\label{finetuning_details}

\textbf{Fine-tuning.} We adopt linear probing with the pre-trained encoder frozen. Models are trained for 100 epochs with a batch size of 128 using AdamW, with a learning rate of $3 \times 10^{-3}$ and weight decay of $1 \times 10^{-4}$. The task-specific heads are defined as follows:
\begin{itemize}[leftmargin=*]
    \item \textbf{Classification:} A linear projection head is used for both single-label (UCM) and multi-label (BigEarthNet) classification, optimized with Cross-Entropy Loss and Multi-Label Soft Margin Loss, respectively.
    
    \item \textbf{Static Urban Indicator Prediction:} A 3-layer MLP regressor predicts GDP and population density from the frozen features using the MSE loss.
    
    \item \textbf{Dynamic Urban Indicator Prediction:} We use a Visual-LSTM architecture. Features from 2010--2020 are projected into a 256-dimensional space and processed by a single-layer LSTM with hidden dimension 256. The prediction is given by
    $\hat{y} = \operatorname{MLP}_{\mathrm{base}}(\mathbf{z}_T) + \operatorname{MLP}_{\mathrm{res}}(\mathbf{h}_{T-1})$,
    where the second term models the temporal development trend.
    
    \item \textbf{Change Detection:} A lightweight U-Net head performs pixel-level change segmentation using multi-stage bi-temporal features. Features are fused by concatenation or absolute difference, followed by upsampling blocks with skip connections and double convolutions, optimized with Binary Cross-Entropy Loss.
\end{itemize}

\section{Temporal Semantic Guidance}
\label{temporal_guidance}
\begin{algorithm}[t]
\caption{Temporal Semantic Guidance}
\begin{algorithmic}[1]
\STATE \textbf{Step 1: Change Extraction}
\STATE \textbf{Requires:} Sequence of satellite images $\{I_1, I_2, \dots, I_T\}$; class labels $\mathcal{C}=\{c_1, c_2, \dots, c_{10}\}$
\FOR{each year $t \in \{1, \dots, T\}$}
    \STATE $S_t \leftarrow \mathcal{G}_{SAM}(I_t)$ \COMMENT{Generate semantic maps for land-use classes $\mathcal{C}$}
\ENDFOR
\FOR{each consecutive pair $(t, t+1)$}
    \STATE $M^{(t, t+1)} \leftarrow \mathcal{A}_{Any}(I_t, I_{t+1})$ \COMMENT{Detect binary change masks at pixel level}
    \FOR{each category $c \in \mathcal{C}$}
        \STATE $\rho_c^{(t, t+1)} \leftarrow \frac{1}{|\Omega|} \sum_{p \in \Omega} (\text{Selector}_c(S_t(p)) \cdot M^{(t, t+1)}(p))$ \COMMENT{Eq. 3: Intersection of change and class}
    \ENDFOR
    \STATE $\vec{\rho}^{(t, t+1)} \leftarrow [\rho_1, \dots, \rho_{|\mathcal{C}|}]$ \COMMENT{Construct yearly semantic change vector}
\ENDFOR

\STATE \textbf{Step 2: Semantic Accumulation}
\STATE Initialize $\vec{\rho}^{(t, t+\delta t)} \leftarrow \vec{0}$
\FOR{$k = t$ \TO $t+\delta t-1$}
    \STATE $\vec{\rho}^{(t, t+\delta t)} \leftarrow \vec{\rho}^{(t, t+\delta t)} + \vec{\rho}^{(k, k+1)}$ \COMMENT{Eq. 4: Linear accumulation of intermediate transitions}
\ENDFOR

\STATE \textbf{Step 3: Semantic Projection and Signal Generation}
\STATE $e^{(t, t+\delta t)} \leftarrow \vec{\rho}^{(t, t+\delta t)} \cdot \mathbf{C}$ \COMMENT{Eq. 5: Weighted aggregation of class embeddings}
\STATE $y \leftarrow 1 - \tanh(\|e^{(t, t+\delta t)}\|_2)$ \COMMENT{Eq. 6: Map change magnitude to soft similarity target}

\RETURN $e^{(t, t+\delta t)}, y$
\end{algorithmic}
\end{algorithm}
The \textit{Temporal Semantic Guidance} organizes the embedding space into interpretable semantic dimensions by leveraging directional urban change signals. The following pseudo code extend the core methodology presented in the main paper and provide additional clarity on how the supervision signal is implemented.

\begin{itemize}[leftmargin=*]
    \item \textbf{Step 1:} We use frozen Grounded-SAM \cite{ren2024grounded} to generate a semantic map $S_t$, assigning each pixel $p$ to one of $|\mathcal{C}|$ urban land-use categories. As defined in Section~3.3, $\operatorname{Selector}_c$ isolates the mask of class $c$. Its pixel-wise product with the binary change mask $M^{(t,t+1)}$, detected by \texttt{Segment Any Change} \cite{zheng2024segment}, captures changes within each functional zone and yields the semantic change vector $\vec{\rho}$.
    
    \item \textbf{Step 2:} Instead of exhaustively comparing all year pairs, we compute yearly transition vectors $\vec{\rho}^{(k,k+1)}$. Changes over any interval $[t,t+\delta t]$ are then obtained by summing category-wise transitions, reducing the complexity from $O(T^2)$ to $O(T)$. This also captures intermediate development stages, such as soil evolving into construction and then tall buildings.
    
    \item \textbf{Step 3:} We embed the 11 urban categories listed in Appendix~\ref{temporal_guidance} using a frozen BERT encoder, producing label embeddings $\mathbf{C}$. The final change embedding $e^{(t,t+\delta t)}$ is computed as a weighted aggregation of these embeddings. The $\tanh$ function in Eq.~6 maps the $L_2$ norm of the semantic shift to a soft similarity target $y \in (0,1)$, encouraging latent visual distances to reflect real-world land-use changes.
\end{itemize}

\section{Baselines}
\label{baselines}
To evaluate the effectiveness of CoST, we compare it against a comprehensive suite of baseline methods, encompassing general-purpose self-supervised learning frameworks and specialized remote sensing (RS) foundation models.
\begin{itemize}[leftmargin=*]
    \item \textbf{SimCLR}~\cite{simclr}:
    A fundamental contrastive learning framework that learns representations by maximizing the agreement between different augmented views of the same data sample.

    \item \textbf{MoCoV3}~\cite{mocov3}:
    An advanced momentum-based contrastive learning approach optimized for Vision Transformers (ViT) to improve training stability and feature quality.

    \item \textbf{DINOv2}~\cite{oquab2023dinov2}:
    A state-of-the-art vision foundation model trained with self-distillation. It generates high-quality, discriminative features that exhibit strong zero-shot performance across a wide range of downstream vision tasks.

    \item \textbf{SeCo}~\cite{manas2021seco}:
    A seasonal contrastive learning pipeline designed for remote sensing, which utilizes temporal revisits of the same geographical location to learn time-invariant representations.

    \item \textbf{CACo}~\cite{mall2023caco}:
    A change-aware contrastive learning method that leverages temporal signals from satellite imagery to distinguish transient seasonal variations.

    \item \textbf{SoftCon}~\cite{wang2024softcon}:
    A soft contrastive learning framework that incorporates multi-label land-cover information.

    \item \textbf{DiNOMC}~\cite{wanyan2024dinomc}:
    An extension of the DINO framework for Earth observation that employs multi-sized local crops and global-local view alignment to capture multi-scale object features.

    \item \textbf{READ}~\cite{read}:
    A lightweight representation learning method tailored for district-level economic scale estimation.

    \item \textbf{PG-SimCLR}~\cite{pgsimclr}:
    A geography-aware model that enhances SimCLR by incorporating Point-of-Interest (POI) data as signals.

    \item \textbf{SatMAE}~\cite{cong2022satmae}:
    A masked autoencoder (MAE) framework specifically adapted for satellite imagery, incorporating temporal and spectral positional encodings to process multi-spectral data.

    \item \textbf{Scale-MAE}~\cite{reed2023scalemae}:
    A scale-aware masked autoencoder that explicitly models  relationship between spatial resolutions by encoding Ground Sample Distance into positional embeddings.
\end{itemize}

\section*{GenAI Usage Disclosure}
We state that AI-assisted technologies were used strictly for language polishing and grammar correction. The authors take full responsibility for the integrity of the work, including the validity of all experimental results and the accuracy of all references.

\bibliographystyle{ACM-Reference-Format}
\bibliography{ref}

\end{document}